\documentclass[10pt,twocolumn]{article} 
\usepackage{simpleConference}
\usepackage{times}
\usepackage{graphicx}
\usepackage{amssymb}
\usepackage{url}
\usepackage{longtable}% for long tables
\usepackage{booktabs}
\usepackage{mathtools}
\usepackage{mathrsfs}
\usepackage{cases}

\DeclareMathOperator{\argmin}{arg\,min}

\begin{document}

\title{Collaboratively Weighting Deep and Classic Representation \\via $l_2$ Regularization for Image Classification}

\author{Shaoning Zeng$^{a,1,2}$, Bob Zhang$^{b,1}$, Yanghao Zhang$^{c,3}$ and Jianping Gou$^{d,4}$ \\
%\\ Technical Report \\
%Seattle, Washington, USA \\
%\today
\\
$^1$ Department of Computer and Information Science, University of Macau, \\
  %Avenida da Universidade, Taipa, Macau, China \\
$^2$ School of Information Science and Technology, Huizhou University, \\
  %Avenue Yanda, NO. 46, Huizhou, Guangdong, China \\
$^3$ Electronics and Computer Science, University of Southampton, \\
  %Southampton SO17 1BJ, United Kingdom \\
$^4$ School of Computer Science and Communication Engineering, Jiangsu University, \\
  %301 Xuefu Road, Zhenjiang, Jiangsu, China \\
%\\ $^a$ zsn@outlook.com,  $^b$ bobzhang@umac.mo,  $^c$ yz16n18@soton.ac.uk, $^d$ goujianping@ujs.edu.cn \\
}

\maketitle
\thispagestyle{empty}

\begin{abstract}
Deep convolutional neural networks provide a powerful feature learning capability for image classification. The deep image features can be utilized to deal with many image understanding tasks like image classification and object recognition. However, the robustness obtained in one dataset can be hardly reproduced in the other domain, which leads to inefficient models far from state-of-the-art. We propose a deep collaborative weight-based classification (DeepCWC) method to resolve this problem, by providing a novel option to fully take advantage of deep features in classic machine learning. It firstly performs the $l_2$-norm based collaborative representation on the original images, as well as the deep features extracted by deep CNN models. Then, two distance vectors, obtained based on the pair of linear representations, are fused together via a novel collaborative weight. This collaborative weight enables deep and classic representations to weigh each other. We observed the complementarity between two representations in a series of experiments on 10 facial and object datasets. The proposed DeepCWC produces very promising classification results, and outperforms many other benchmark methods, especially the ones claimed for Fashion-MNIST. The code is going to be published in our public repository\footnote{https://github.com/zengsn/research}. 
\end{abstract}

\section{Introduction}

Machine learning methods have been applied to deal with various multi-media and computer vision tasks. Traditionally, linear models such as sparse representation (SR) \cite{Wright2009Robust} and collaborative representation (CR) \cite{Zhang2012Sparse} have drawn much attention and gained promising results in image classification. Lately, nonlinear deep learning \cite{LeCun2015Deep} models, e.g., ResNet \cite{He2016Deep} and VGG \cite{Simonyan2014Very}, have produced state-of-the-art results in many image-based tasks, including face recognition, object detection, video tracking, etc. 
Linear sparse models can be utilized to improve deep neural networks \cite{Xin2016Maximal}. On the other hand, more and more conventional methods took deep features as input to gain more promising classification results \cite{Cai2016Probabilistic,Zeng2017Joint,Zeng2018Robust}. However, recent studies showed that deep features from neural networks are usually designed for SVM-like classifiers \cite{Li2018Decision}. Using deep features as sole input in non-SVM classical models could be dubious. For this problem, we believe that the technique of linearly representing images can be applied to enhance nonlinear deep models.  

Deep learning shows a very strong capacity to learn discriminative image features. CNN features off-the-shelf \cite{Sharif2014Cnn} were demonstrated to be powerful for recognition tasks. Learning deep features can help to obtain state-of-the-art results for different tasks like face recognition \cite{Wen2016Discriminative}, scene recognition \cite{Zhou2014Learning}, person re-identification \cite{Xiao2016Learning} and general image classification \cite{Shao2017Performance,Valada2018Deep}. The good news is that many conventional machine learning methods can also learn credible features from images. In recent years, Sparse Representation (SR) \cite{Wright2009Robust} via $l_1$ regularization has shown huge potential in feature extraction and image classification. On the other hand, $l_2$ regularization-based Collaborative Representation (CR) \cite{Zhang2012Sparse} can also build a similarly robust linear model. The $l_2$ regularization inside CR helps to create an equally discriminative but faster sparse representation \cite{Xu2017New}. According to our observation, sparseness plays an important role in both linear and nonlinear models. It is likely for these two paradigms, deep and classic representations, to generate a new representation learning model when collaborating with each other.       

In this paper, we propose a Collaborative Weight-based Classification method that brings deep and classic non-deep representation together, to implement a more promising image classification. We name it DeepCWC for short. The contribution of this work includes: 1) proposing a new classifier to integrate features from linear and nonlinear models, 2) giving an analysis on how black-box deep features work in a sparse classification model, 3) conducting image classification experiments on different CNN models and convolutional layers inside them, to demonstrate the performance of DeepCWC in a consistent and comprehensive way. The proposed method produces promising results on face and object recognition. In particular, it ranks first in recognition (97.66\%) on the Fashion-MNIST dataset.

\section{Related Work}
\label{Idea}

\begin{figure*}[htp]
\begin{center}
\includegraphics[width=\textwidth]{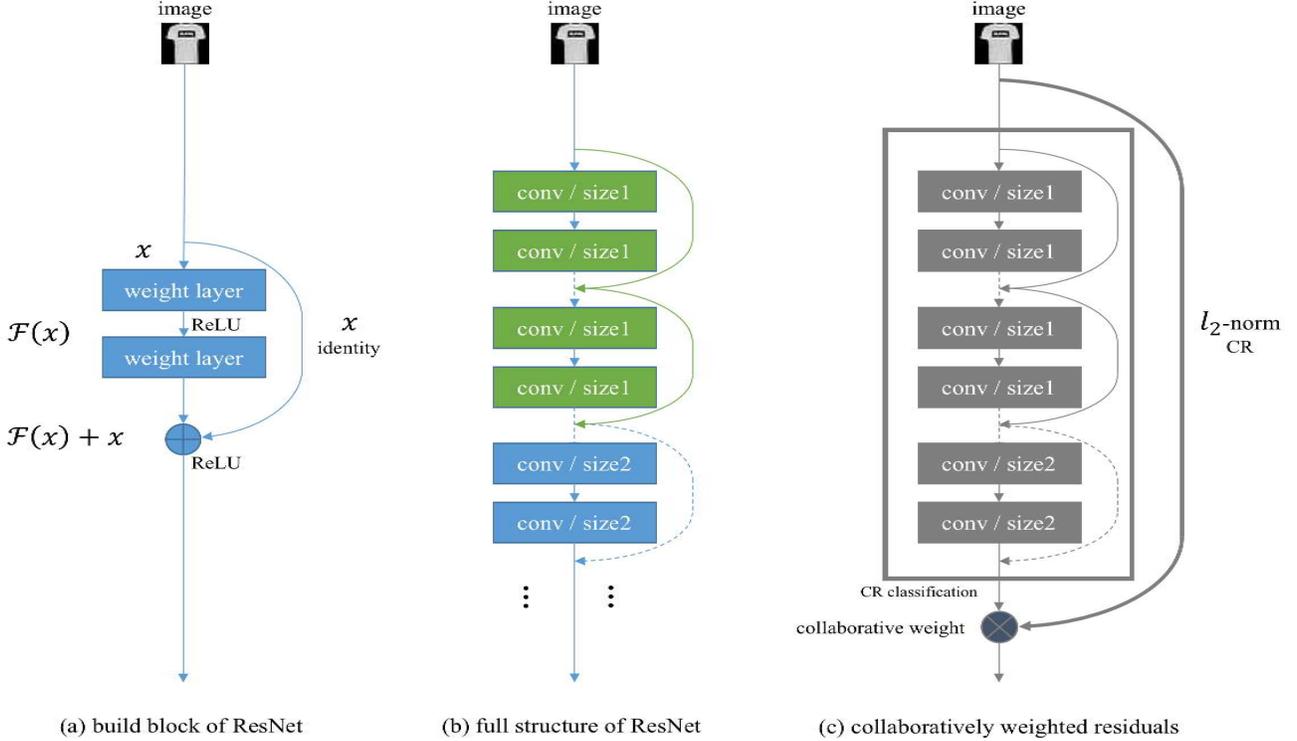}
\end{center}
\caption{Collaboratively weighted residuals of CR and ResNet models.}
\label{fig:DeepCWC} 
\end{figure*}

The root inspiration comes from the popular deep residual network (ResNet) \cite{He2016Deep}. ResNet keeps an identity map learned from the last layer, and applies it to next layer of learning. Then, it constructs a new building block $y = \mathscr{F}(x,\{W_i\}) + x$, as shown in Fig. \ref{fig:DeepCWC}(a). This explicitly allows these layers to fit a residual mapping, so as to make it easier for the residual to be zero (sparse) than to fit an identity mapping by a stack of nonlinear layers. In this way, the discrimination learned in the previous layers will be propagated layer-by-layer. Also, there would be a linear transformation between some layers, if two connected blocks have a different dimension, as shown in Fig. \ref{fig:DeepCWC}(b). Denote the linear projection as $W_s$, where the building blocks become $y = \mathscr{F}(x,\{W_i\}) + W_sx$. 

ResNet attracts great attention and progresses observably \cite{He2016Identity}, while our main focus is the way how it utilizes the prior information. It is possible to include sparse learning, as prior information, in deep neural networks as well. For example, grouping multiple sparse regularizations for simultaneously optimizing deep neural networks \cite{Scardapane2017Group}. Wen et al. proposed to learn a structured sparsity in deep neural networks to regularize the inside structures (i.e., filters, channels, filter shapes, and layer depth) \cite{Wen2016Learning}. Afterwards, a fixed linear sparse filter can be cascaded with a thresholding nonlinearity to maximize sparsity in deep neural networks \cite{Xin2016Maximal}. It becomes an emerging trend to utilize linear sparse models to collaborate with nonlinear neural networks.  

As shown in Fig. \ref{fig:DeepCWC}(c), our idea has a similar structure following the building block of ResNet. The key is fusing the identity map learned from $l_2$-norm collaborative representation \cite{Zhang2012Sparse} to the result after deep residual learning. To simplify the implementation structure, the linear model is not injected into the building block of the neural network. Instead, it performs on the classifier. There are several reasons for this structure. Firstly, it helps to avoid overheads in the training process of the neural networks. Furthermore, it creates a more general structure that can be easily extended to other types of neural networks, which are not limited to ResNet. For example, we also implement this in Inception \cite{Szegedy2017Inception} and VGG \cite{Simonyan2014Very}, which will be demonstrated in Sec. \ref{Results}. The idea behind pairing nonlinear deep learning with an additional linear representation is to make the network more capable in different classification tasks. 

However, the usage of the prior learned information is different in our implementation. ResNet adds up the learned $x$ in model training, while the proposed DeepCWC will introduce a collaborative weight in classification, which is obtained by an element-wise multiplication instead of addition. The next section explains the detailed implementation.

\section{Deep collaborative weight-based classification}
\label{DeepCWC}

The key idea in Deep Collaborative Weight-based Classification is straightforward: using the model of Collaborative Representation (CR) \cite{Zhang2012Sparse} to learn a classifier from the original images and deep features in pairs. CR is based on $l_2$ normalization and emphasizes the collaboration among all samples in the representation. However, more and more evidence points to the fact that the collaboration requires help from the sparseness in the representation to maintain a high level of performance \cite{Akhtar2017Efficient}. We believe that using deep features is one of the possible solutions. 

\subsection{Pair of residual learning}
\label{PairLearning}

In CR based classification (CRC), the role of collaboration among classes is stressed, rather than sparsity in the representation, when representing a test sample. Let $A$ denote the training samples selected from all $C$ classes, while $y$ is the test sample. Both $A$ and $y$ will be normalized to have $l_2$-norm. The representation of $y$ by $A$ can be denoted as an approximate linear problem $y \approx A \alpha$, where $\alpha = [\alpha_1,\alpha_2,\dots,\alpha_C]$ is the representation coefficient to be solved.

First of all, a regularized least square method \cite{Zeng2018Robust} is used to solve the problem and perform the collaborative representation of the original image sample as follows
\begin{equation} \label{eq:crc_img_coef}	
	(\hat \alpha)= \argmin_{\alpha} \{\big\|y - A\cdot\alpha\big\|_2^2\ + \lambda\ \big\|\alpha\big\|_2^2 \}, 
\end{equation}
where $\lambda$ is the regularization parameter, which introduces a certain number of ``sparsity'' to the solution. The solution of this linear problem by using regularized least square can be derived as
\begin{equation} \label{eq:crc_img_coef}	
	\hat{\alpha}= (A^TA + \lambda \cdot I)^{-1} A^Ty.
\end{equation}
Let $P = (A^TA + \lambda \cdot I)^{-1} A^T$, such that $P$ is a projection matrix that can be pre-solved and independent of the test sample $y$. The projection makes CR much faster than the conventional SR. It is noted that this operation may not fit in the memory of a large-scale dataset \cite{Yu2012Large}. In our implementation, we used an incremental strategy \cite{Xiao2014Error,Ristin2014Incremental} to deal with this problem, despite the fact that there are other potential solutions, e.g., dictionary learning \cite{Cai2016Probabilistic,Xu2017Sample}.  

From this, we can obtain the coefficient vector $\hat{\alpha_i}$ related with the $i$th class. Typically, in SRC the coefficient is utilized to solve the representation residual of a specific class by $\big{||}y - A_i \cdot \hat{\alpha_i} \big{||}_2$. Besides this, in the original CRC implementation, the $l_2$-norm of the sparse coefficient $\big{||}\hat{\alpha_i} \big{||}_2$ was also added to obtain more discrimination when performing classification \cite{Zhang2012Sparse}. Finally, the residual is obtained by
\begin{equation} \label{eq:crc_img_res}	
	res_{img} = \big{||}y - A_i \cdot \hat{\alpha_i} \big{||}_2. % + \lambda\ \big\|\alpha_i\big\|_2.
\end{equation}

%\subsection{Deep features extraction}
%\label{Deep_Features}

At the same time, a side-by-side CR is performed on deep features. The so-called deep features are specific to the layer in the deep model. Theoretically, any layer can be utilized to extract deep features. For example, the $global\_pool$ layer \cite{He2016Deep,He2016Identity,Szegedy2017Inception} in the ResNet and Inception models.
%In deep transfer learning, we can reuse the features described on all layers of the deep neural network except the last layer, since the last layer in fact does not contain any features but servers as a classifier or classification layer. When transferring the deep model to another different domain, we are likely to switch to a different well-chosen classifier. Before this, we extract the deep features by using a deep model, for example, ResNet, VGG, Inception, etc. 
Here, we denote the deep features from one specific layer ($j$) of the neural networks for all training samples as 
\begin{equation} \label{eq:crc_b_fea}	
	B = \mathit{fea}_{cnn}(A, layer_{j}),
\end{equation}
and therefore the query sample becomes  
\begin{equation} \label{eq:crc_y_fea}	
	y_{cnn} = \mathit{fea}_{cnn}(y, layer_{j}).
\end{equation}
%All deep features are obtained from the layer before the last one in a deep neural network. Normally, the feature set obtained with a deep neural network has a different size from the size of the original samples. For example, the size of features extracted by the ResNet is 1000, no matter what size the original images are. 
The size of the feature set is determined by the design of the specific layer, where it is normally mismatched with the size of the original images. It is hard to simply perform integration on the feature level. Therefore, fusion of the feature pairs will be performed on the residuals, which have the size same as the class number.

%\subsection{CR of the deep features}
%\label{CR_Features}
%When the last layer of features in a deep neural network is obtained, they can be used as samples to perform classification, which typically happens in the last layer in the deep neural network. In performing deep transfer learning, we utilize CR instead of the original classifier in the deep model, e.g. SVM, to classify the sample. 
Then, the representation coefficient obtained from the deep features by a similar CR process is
\begin{equation} \label{eq:crc_deep_coef}	
	\hat{\beta}= (B^TB + \lambda \cdot I)^{-1} B^T y_{cnn}.
\end{equation}

After that, the residual between the query feature set $y_{cnn}$ and each class of training feature sets can be solved with the same method as Eq. \eqref{eq:crc_img_res}
\begin{equation} \label{eq:crc_deep_res}	
	res_{cnn} = \big{||}y - B_i \cdot \hat{\beta_i} \big{||}_2. % + \lambda\ \big\|\beta_i\big\|_2.
\end{equation}

%In theory, deep learning can output a more promising classification than the conventional methods, because the deep features contain a more discriminative feature set. This is true in all deep learning models, which is demonstrated in our experiments. However, it is not always true when using a pre-trained model. No matter how stable a deep model performs in a domain, it might be unstable when applied to a different domain. We demonstrated this phenomenon in our experiments as well, and we believe that this is likely to be caused by missing some important information.  

\subsection{Fusion based on collaboratively weighting}
\label{Fusion}

Our proposed method manages to retrieve this part of the missed information via a novel fusion operation. The fusion is performed on two residuals, since both have an equal dimension depending on the number of classes. %However, the representation coefficients may not have the same size. For example, whatever size the image samples are, the deep features are always in a consistent size of 1000 when obtained by using the ResNet model. What is more, the number of classes is often much smaller than this. 
Therefore, fusion on the residuals is not only straightforward, but also faster. 

Let us denote the residuals solved from two groups of samples as \\$res_{img} = [d_{img,1}, d_{img,2}, \dots, d_{img,C}]$ and $res_{cnn} = [d_{cnn,1}, d_{cnn,2}, \dots, d_{cnn,C}]$, where $C$ is the number of classes in the dataset. Then, the fusion via the collaborative weight is performed on the residual vector via an element-wise multiplication, 
\begin{equation} \label{eq:cwc_res}	
	%d_{fusion,i} = d_{cnn,i} * \frac{d_{img,i}}{max([d_{img,1}, \dots, d_{img,C}])}.
	res_{fusion} = res_{img} \odot res_{cnn}, %\\ 
	%= [d_{img,1}, d_{img,2}, \dots, d_{img,C}] \odot [d_{cnn,1}, d_{cnn,2}, \dots, d_{cnn,C}], 
\end{equation}
where the residual entry related with the $i$th class is calculated by the collaborative weight. This weight means that each entry in the residual vector is assigned a weight solved by the collaborative representation of the original images. The information carried by this weight compensates the missing part of the abstract higher layers in a neural network. In this way, we obtain the final fusion residual. Although additional or weighted averages are a more common approach to perform fusion in many other methods \cite{Zeng2017Improving,Zeng2017Joint,Wen2018Adaptive}, they require a set of fine-tuned factors to obtain a good result. What is more, we observed a descending accuracy when adding up two residuals. 

%\begin{equation} \label{eq:cwc_weight}	
%	res_{fusion} = [d_{fusion,1}, d_{fusion,2}, \dots, d_{fusion,C}].
%\end{equation}

Finally, we classify the test sample to a class with minimal residual as follows
\begin{equation} \label{eq:cwc_identity}	
	identity(y) = \argmin_i(res_{fusion,i}).
\end{equation}

The idea of our collaborative weight is simplistic and intuitive. The collaborative weights are determined by the relative contribution of each class from the original samples. Each residual of the deep features is overlapped by a weight solved using the collaborative representation of the original samples, in order to integrate its contribution. 

\subsection{Why deep features works in CR}
\label{WhyDeep}

\begin{figure*}[htp]
\begin{center}
\includegraphics[width=14cm,height=7cm]{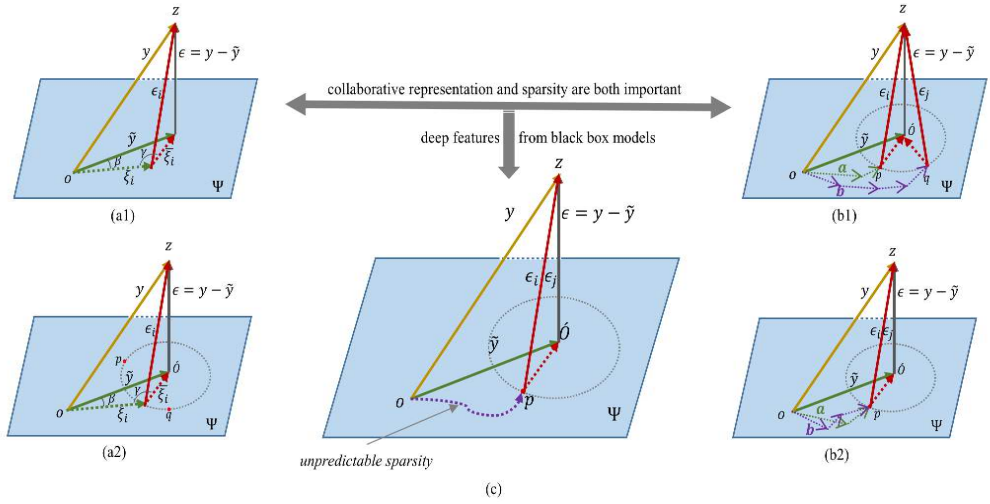}
\end{center}
\caption{The sparsity from deep features in the CR model.}
\label{fig:SparsityCR} 
\end{figure*}

To answer this question, we first need to answer another question: What is the relationship between collaboration and sparsity? Many had tried to give an answer with some considering sparsity as being more important \cite{Wright2009Robust,Deng2013Defense}. On the other hand, others insist that collaboration matters more \cite{Zhang2012Sparse,Peng2014Learning}. However, as for the rest, they treat collaboration and sparsity as equal factors \cite{Zeng2017Multiplication,Zeng2017Improving,Akhtar2017Efficient}. So far, the last viewpoint well explains our proposed collaboratively weighting deep and non-deep representation.  

The subspace occupied by the columns of a sparse dictionary $\Phi$ can be denoted as a set of $\Psi$. Fig. \ref{fig:SparsityCR} shows the geometrical illustration of this subspace. A test sample $y$ can be approximately represented by the columns of $\Phi$, and the error is $\epsilon = y - \tilde y$. In addition, vector $\tilde y$ can be decomposed to $\xi_i$ and $\tilde \xi_i$, as depicted in Fig. \ref{fig:SparsityCR} (a1). According to \cite{Zhang2012Sparse}, the angles $\beta$ and $\gamma$ together decide the robustness of the CR model. However,  Fig. \ref{fig:SparsityCR} (a2) shows that it is likely to have more than one right answer, which is depicted as the circle. The distances of $pz$ and $qz$ are the same. Therefore, CR by itself without considering sparsity may not be robust enough \cite{Akhtar2017Efficient}.

Fig. \ref{fig:SparsityCR} (b1) and (b2) show how the sparsity can help in the CR model. In these two cases, where (b1) $\xi_i \neq \tilde \xi_i$ and (b2) $\xi_i = \tilde \xi_i$, $a$ and $b$ are two paths to points $p$ and $q$. The distances are also the same $||\epsilon_i||_2 = ||\epsilon_j||_2$ in these two instances, while $a \neq b$ in (b1) and $a = b$ in (b2). This means that the class-specific residuals are equal in both cases, but construction of vectors $\xi_i$ and $\xi_j$ may be different. The components consisting of the path depict the sparsity in the representation, where fewer steps of $\xi_i$ indicates a sparser collaborative representation coefficient $\alpha$. Using sparser features in CR can help to produce a more robust classification, which is the very reason why DeepCWC works. 

The black box deep model provides an unpredictable sparsity in the deep features. Currently, we can only accept this fact according to the largely contracted dimension of the deep feature set, with feature learning being the most powerful characteristic of deep learning. As shown in Fig. \ref{fig:SparsityCR} (c), the effort of CR on deep features can be painted as a random and unpredictable curve between $o$ and $p$. This can be treated as potentially the most efficient path and is also the result observed in the experiments.     

\subsection{Why the fusion is positive}
\label{WhyFusion}

\begin{figure*}[htp]
\begin{center}
\includegraphics[width=\textwidth]{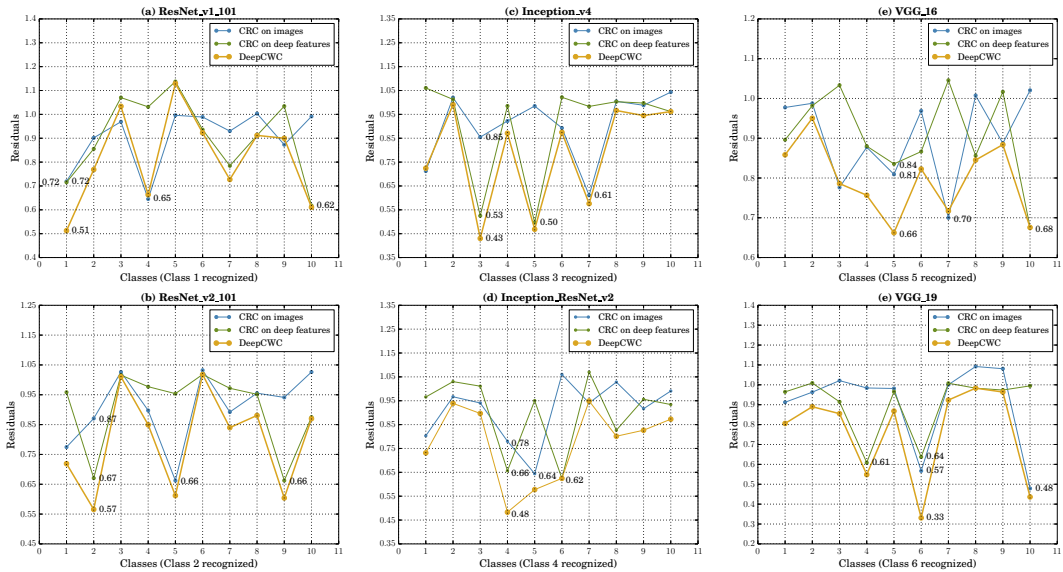}
\end{center}
\caption{The residuals from the pair of CR representation and the final fusion.}
\label{fig:Residuals} 
\end{figure*}

When the deep features are ready and fit well in the CR model, the next problem is how to consolidate both of them into an united set. This is where the collaborative weight works. Previous work showed that well constructing the residuals is helpful to generate a robust sparse model, i.e., the two-phase sparse representation model \cite{Xu2011Two}. To illustrate the impact of the weight, we captured some runtime data from our experiments, which is plotted in Fig. \ref{fig:Residuals}.

The purpose of the collaborative weight is to expand the more promising residuals, while restricting the other ones. The target class is selected by a final minimization, hence, we look for the smallest values. For example, the residuals of classes 1-10 are shown on the $ResNet\_v1\_101$ model in Fig. \ref{fig:Residuals} (a). The correct class label is 1, where the distances of CRC on images and deep features are both below 1 (0.72). However, the minimal values of them are 0.65 and 0.62, respectively. This could lead to a wrong classification result (Class 4 and 10). After the fusion, the resultant residual becomes even smaller, resulting in the correct class being chosen (Class 1). This ensures that the classification will not be affected by other nearby classes. The same phenomenon can be found on other models, which are annotated in Fig. \ref{fig:Residuals} (b) - (f). On the other hand, the classes with a larger distance value, e.g., $d_i > 1.0$, the fusion will make it much larger ($m * n > 1.0$, if $m>1.0$ and $n>1.0$). This in turn helps to avoid negative results. 

Another point to note in the fusion is that the parameters do not need to be tuned. This is not like other conventional fusion schemas, which usually introduces one or two fusion factors \cite{Zeng2017Improving,Zeng2017Joint,Wen2018Adaptive}, where more parameters call for more tuning. The fusion result is determined by the residuals themselves. We barely need to look for an optimal parameter and maintains the effectiveness of DeepCWC. 
%------------------------------------------------------------------------
\section{Experimental results}
\label{Results}

This section describes the experiments to demonstrate the robustness and performance of the proposed method. First of all, six facial datasets, including FERET \cite{Phillips2000Feret}, MUCT \cite{Milborrow10MUCT}, Yale B \cite{Georghiades2001FewYaleB}, Georgia Tech (GT) \cite{Chen2005FaceGT}, AR \cite{Martinez1998AR}, and ORL \cite{OrlData}, are selected to evaluate the performance of face recognition. These experiments were conducted due to the fact that CRC is usually applied to face recognition.  Secondly, another set of experiments had been performed on some object datasets, including a leaf dataset Flavia [26], and three object datasets CIFAR-10 \cite{Krizhevsky2009LearningCIFAR10}, Fashion-MNIST \cite{Xiao2017Fashion} and COIL-100 \cite{Nayar1996Columbia}, which are often utilized to evaluate deep learning methods. 

Also, in order to evaluate the robustness after introducing the collaborative weight between the original images and the deep features, we extracted the deep features using multiple state-of-the-art deep CNN models, including ResNet\_v1\_101, ResNet\_v2\_101, Inception\_v4, Inception\_ResNet\_v2, VGG\_16 and VGG\_19. All of these are trained previously on the ImageNet dataset \cite{Deng2009Imagenet} in Google TensorFlow\footnote{https://github.com/tensorflow/models/tree/master/research/slim\#Pretrained}. Our assumption is the proposed DeepCWC works on different deep CNN models. The feature extraction is performed on the TensorFlow-Slim library. Besides this, another goal is to investigate which layer of features in a CNN model are more suitable for collaborative weight. Based on the considerations, we conducted a set of relevant experiments and obtained the following results.
 
\subsection{Experiments for face recognition}
\label{FaceRecognition}

\begin{figure*}[htp]
\includegraphics[width=\textwidth]{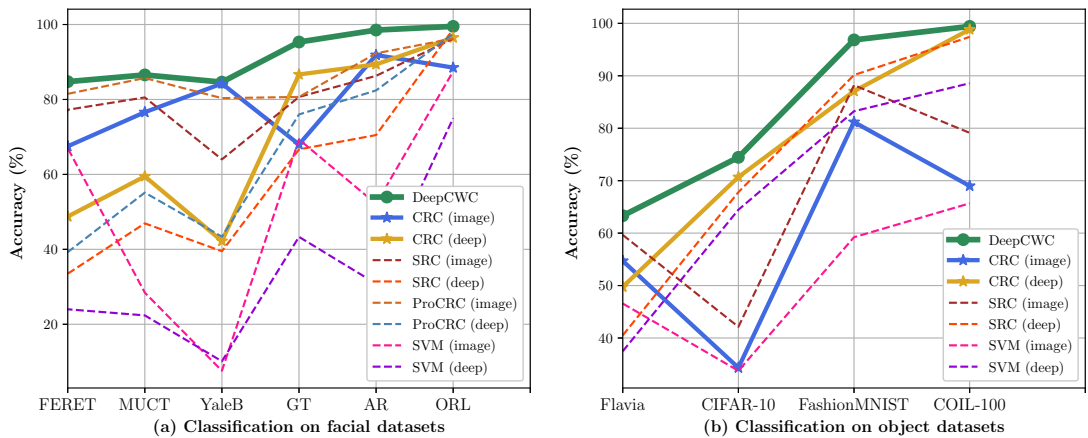}
\caption{Face and object recognition accuracy comparison.}
\label{fig:accFaceObjRecResNet1} 
\end{figure*}

We ran experiments on six popular benchmark facial datasets. These datasets are relatively small. The smaller datasets do not contain enough samples to train a robust model by CNN, but we can extract the deep features using pre-trained deep CNN models. Our goal in this group of experiments is to compare the classification result between our proposed method and state-of-the-art methods. The best results are shown in Fig. \ref{fig:accFaceObjRecResNet1} (a).

It is clear that the proposed DeepCWC method outputs a higher recognition accuracy than normal CRC, CRC using deep features and other state-of-the-art methods, no matter which CNN model is used to extract the deep features. It is uncertain that using deep features would generate a higher recognition accuracy than using the original images. For example, when utilizing the ResNet\_v1\_101 model to extract features of the AR dataset, CRC performs better on original images than deep features, as shown in Fig. \ref{fig:accFaceObjRecResNet1} (a). And this is also true in some other experimental cases, which shows one of the limitations of a typical deep learning method. This is the very reason why we proposed the DeepCWC.

No matter which dataset, performing fusion of two feature sets based on the collaborative weight generates a higher recognition accuracy. Even in cases that merely use deep features without collaborative weight, e.g., on FERET, MUCT, Yale B, GT, AR and ORL, the proposed DeepCWC helps the recognition by fusing features from the original images and the CNN models. 

\subsection{Experiments for object recognition}
\label{ObjectRecognition}

The next set of experiments were performed on some leaf and object datasets, including Flavia (leaf), COIL-100, CIFAR, and Fashion-MNIST. The results are consistent on all datasets, as shown in Fig. \ref{fig:accFaceObjRecResNet1} (b). Incorporating the deep features learned by ResNet\_v1\_101, the recognition accuracy (yellow) is much higher than the result on the original images (blue), except the result on the Flavia. However, DeepCWC further pushes the recognition up to an even higher level, and the improvements are stable on all datasets.   

The highest accuracy is obtained on the COIL-100 dataset when using the first 60 samples in each class (83\%) as the training samples. Deep features are beneficial to classification on this dataset, where DeepCRC (up to 98.83\%) outperformed CRC (only 69.0\%) by over 30\%. That being said, the proposed DeepCWC still produces the highest accuracy of 99.42\%, which reached a state-of-the-art level in recognition. On the Flavia leaf dataset, the improvement generated by collaborative weight is remarkable, though the accuracy is relatively lower, as shown in Fig. \ref{fig:accFaceObjRecResNet1} (b). The results on the CIFAR-10 dataset get an improvement as well. And the improvement (the column Impr) is calculated by the rate of the accuracy from the DeepCWC over the higher one between CRC on images and deep features, and the improvements on the Flavia and Fashion-MNIST are up to 21.01\% and 12.41\%, respectively.

\subsection{Experiments on different layers}
\label{DeepLayers}

\begin{figure*}[hbp]
\includegraphics[width=\textwidth]{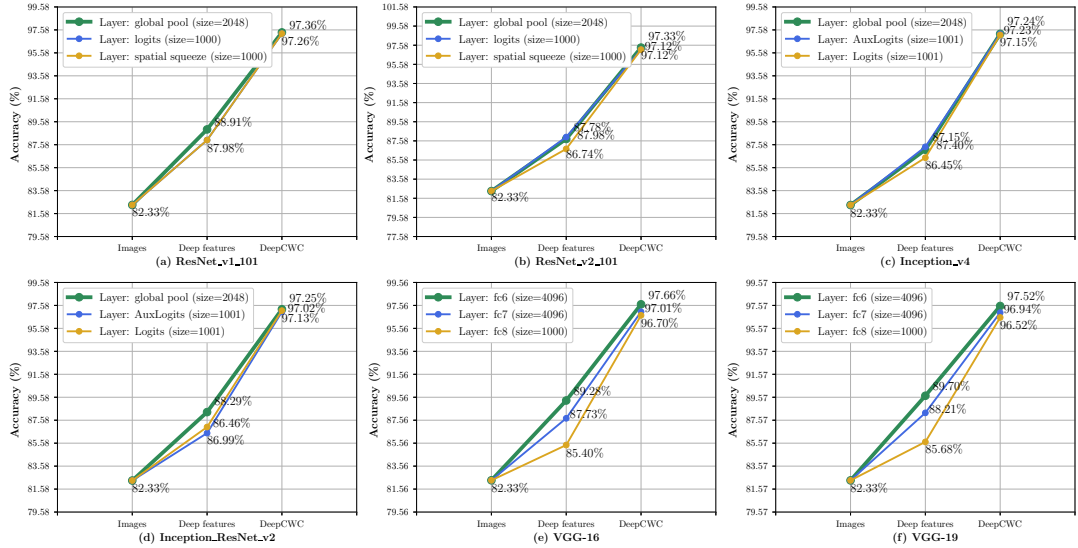}
\caption{Accuracies with different layers of six deep models.}
\label{fig:DeepLayers} 
\end{figure*}

Two versions of the ResNet pre-trained models, ResNet\_v1\_101 \cite{He2016Deep} and ResNet\_v2\_101 \cite{He2016Identity}, are tested in this set of experiments. As described above, we borrowed a similar architecture idea from the deep residual network, as shown in Fig. \ref{fig:DeepCWC}. For this reason, we design the first implementation of DeepCWC based on ResNet. There are 101 layers in the network, and we evaluate the proposed method on the features obtained from three layers, which are $global\_pool$, $logits$\footnote{$resnet\_v1\_101/logits$ or $resnet\_v2\_101/logits$.} and $spatial\_squeeze$\footnote{$resnet\_v1\_101/spatial\_squeeze$ or $resnet\_v2\_101/spatial\_squeeze$.} from shallower to deeper. The layer $spatial\_squeeze$ is the layer before the last convolutional layer, while the $logits$ layer is before $spatial\_squeeze$. These two layers have the same feature set size (1000). However, the $global\_pool$ layer is before this layer and has a larger size of 2048. The classification results are demonstrated in Fig. \ref{fig:DeepLayers} (a) and (b).

In both models, performing CRC on deep features obtained from each layer produces a higher accuracy than using the original images. DeepCWC generates an even higher accuracy than both of these. We observe exactly the same result ($97.26\%$) in both $logits$ and $spatial\_squeeze$ layers. For classification, the feature maps of the last convolutional layer are fed into fully connected layers followed by a softmax logistic regression layer \cite{Krizhevsky2012Imagenet}. The global average pooling \cite{Lin2013Network} is introduced to avoid overfitting in the fully connected layers. Using the features maps captured from the $global\_pool$ layer produces a slightly higher accuracy ($97.36\%$). Every result reaches a state-of-the-art level, and is higher than all current implementations (See subsection \ref{StateOfTheArt}). It is noted that the computation time increases due to a larger size (double) of the feature maps from layer $global\_pool$. Therefore, the $logits$ (or $spatial\_squeeze$) layer should be a better choice when considering the balance between accuracy and speed. 

To investigate the performance when using a different CNN model, two Inception models are evaluated in a similar way. They are Inception\_v4 and Inception\_ResNet\_v2 models pre-trained on ImageNet. Besides the $global\_pool$ layer, the $Logits$ and $AuxLogits$ layers are also utilized to extract deep features, before being fed into to the linear CR model. A set of similar results are observed in the experiments, as shown in Fig. \ref{fig:DeepLayers} (c) and (d).

DeepCWC achieves an accuracy over $97\%$ on deep features obtained from three layers. The highest result ($97.24\%$) is the one with the largest size (2048) from the $global\_pool$ layer. In this case, the $AuxLogits$ layer, with a smaller size (1001) than $global\_pool$, produced an approximately equal accuracy of $97.23\%$. The result from $Logits$ is close to this. In fact, all of the results in DeepCWC for the three cases are stable and close to each other. 

The last set of models are of the VGG implementation. We chose VGG-16 and VGG-19 models, and utilized the feature maps from their $fc6$, $fc7$ and $fc8$ layers. The size of both $fc6$ and $fc7$ is 4096, while the last $fc8$ layer has a smaller size of 1000. The largest feature set in this group of experiments produced the most promising classification results. What is more, the trend is consistent with before. As shown in Fig. \ref{fig:DeepLayers} (e) and (f), the highest accuracy is up to 97.66\%, using the shallower $fc6$ layer in VGG-16, which is also the most promising result we obtained using this dataset and in all cases. The results achieved by VGG-19 are slightly lower than this, but higher than the other cases. The larger feature size helps to produce a more accurate classification. 

\begin{table*}
  \caption{State-of-the-art accuracies on the Fashion-MNIST dataset.}
  \label{table:StateOfArt}
  \center
  \begin{tabular}{llr}
    \toprule
    Deep Model	($\ast$)				& Preprocessing 	& Highest Accuracy  \\
    \midrule
    CapsNet      						& None 			& 90.6\% \\
    VGG16 26M parameters 			& None 			& 93.5\% \\ 
    GoogleNet with cross-entropy loss 	& None 			& 93.7\% \\ 
    %ResNet18 						& Yes 			& 94.9\% \\ 
    MobileNet 						& Yes   			& 95.0\% \\ 
    DenseNet-BC 768K params 			& Yes   			& 95.4\% \\ 
    Dual path wide resnet 28-10 			& Yes   			& 95.7\% \\ 
    %WRN-28-10 						& Yes   			& 95.9\% \\ 
    WRN-28-10 + Random Erasing 		& Yes   			& 96.3\% \\ 
    WRN40-4 8.9M params 			& Yes   			& 96.7\% \\ \hline
    Ours  							&    				&  \\ \hline
    DeepCWC with Inception\_v4		& None (global pool) & 97.24\% \\
    DeepCWC with Inception\_RN\_v2	& Yes (global pool) 	& 97.25\% \\
    DeepCWC with ResNet\_v2\_101		& Yes (global pool)   	& 97.33\% \\
    DeepCWC with ResNet\_v1\_101		& None (global pool)  & 97.36\% \\ 
    DeepCWC with VGG\_19			& None (fc6)  		& 97.52\% \\
    DeepCWC with VGG\_16			& None (fc6)   		& \textbf{97.66}\% \\ \hline
    \multicolumn{3}{l}{$\ast$ Data claimed in https://github.com/zalandoresearch/fashion-mnist  } \\
    \multicolumn{3}{l}{$\ast$ CapsNet in https://github.com/naturomics/CapsNet-Tensorflow\#results  } \\
   % \multicolumn{3}{l}{$\star$ (gp) refers to the $global\_pool$ layer.  }
  %\bottomrule
\end{tabular}
\end{table*}

\begin{table*}
  \caption{Classification time and speed on different models.}
  \label{table:time}
  \center
  \begin{tabular}{llcrr}
    \toprule
    Model			& Layer 			& Feature Size 	& Time (sec) 		& Speed (sec/image)  \\
    \midrule
    VGG-16		& /fc6   			& 4096	& 16342.347	& 0.233 \\
    %VGG-16		& /fc7   			& 4096	& 16531.621	& 0.236 \\
    VGG-16		& /fc8   			& 1000	& 6826.632 	& 0.098 \\
    VGG-19		& /fc6   			& 4096	& 16567.730 	& 0.237 \\
    %VGG-19		& /fc7   			& 4096	& 16460.812 	& 0.235 \\
    VGG-19		& /fc8   			& 1000	& 6526.632 	& 0.093 \\
    Inception\_v4	& /Logits   		& 1001 	& 7177.612	& 0.103 \\
    %Inception\_v4	& /AuxLogits   		& 1001 	& 7193.574	& 0.103 \\
    Inception\_v4	& /global\_pool   	& 1536 	& 8626.181	& 0.123 \\
    Inception\_RN\_v2	& /Logits   		& 1001 	& 6734.609	& 0.096 \\
    %Inception\_RN\_v2	& /AuxLogits   		& 1001 	& 6839.672	& 0.098 \\
    Inception\_RN\_v2	& /global\_pool   	& 1536 	& 7884.362	& 0.113 \\
    ResNet\_v1\_101	& /global\_pool   	& 2048 	& 9108.256	& 0.207 \\
    ResNet\_v1\_101	& /logits   			& 1000 	& 7302.812	& 0.104 \\
    %ResNet\_v1\_101	& /spatial\_squeeze  & 1000 	& 7487.812	& 0.107 \\ 
    ResNet\_v2\_101	& /global\_pool   	& 2048 	& 8373.421	& 0.120 \\
    ResNet\_v2\_101	& /logits   			& 1000 	& 7254.657	& 0.104 \\ \hline
    %ResNet\_v2\_101	& /spatial\_squeeze  	& 1000 	& 7110.983	& 0.102 \\ \hline
  %\bottomrule
\end{tabular}
\end{table*}

\subsection{Comparison to the state-of-the-arts on Fashion-MNIST}
\label{StateOfTheArt}

The results obtained on the pre-trained models (over 97\%) are all state-of-the-art, as shown in in Tab. \ref{table:StateOfArt}. According to the description of the current methods, all of them are tuned and trained on the Fashion-MNIST dataset locally. Also, most of them applied one or two preprocessing techniques, and used the same deep neutral network architecture, e.g., VGG, ResNet, etc. Previously, the most promising accuracy was obtained by the Wide Residual Networks (WRN) model \cite{Zagoruyko2016Wide}, both of which applied the standard preprocessing (mean/std subtraction/division) and augmentation (random crops/horizontal flips). For example, one used the random erasing technique \cite{Zhong2017Random} and produced an accuracy of 96.3\% \footnote{https://github.com/zhunzhong07/Random-Erasing}, while the other one with 96.7\% accuracy had 8.9 M parameters and utilized freezing layers \footnote{https://github.com/ajbrock/FreezeOut}. 

Compared to current state-of-the-art methods, the proposed DeepCWC produces a higher result using multiple CNN models. The classification accuracy ranges from 97.24\% to 97.66\%, which are all higher than previous methods. The highest accuracy is generated on VGG-16 from the $fc6$ layer with a size of 4096. 

\subsection{Discussion}
\label{Discussion}
%\subsection{Computation time}
%\label{Time}

Our experimental machine was configured with the following hardware, including an Intel$\textsuperscript{\circledR}$ Core$\textsuperscript{TM}$ i7-7820X CPU@3.60GHz x 16, 64 GB RAM, 1.3 TB SSD and one NVIDIA TITAN Xp GPU. The code was run on TensorFlow 1.6, MATLAB R2016 and Ubuntu 16.04 OS. The recorded time consumption of each experimental case is shown in Tab. \ref{table:time}.

This time includes the whole training and testing of both the original images and deep features in CR, but does not count the time for feature extraction by the pre-trained models. Therefore, the speed (seconds per sample) is calculated by dividing the total time by the size of dataset (70000). Considering that the running state of the machine may fluctuate, the speed is between 0.1 - 0.2 seconds per sample. Furthermore, the following can be discussed about the proposed method.   

\textbf{Linear representation such as CR can improve deep neural networks based representation learning}. Even using a pre-trained model, the proposed DeepCWC achieved a state-of-the-art classification result on the Fashion-MNIST dataset. The accuracy and performance outperformed current popular methods as well. This gives us a clue that linear methods have a new way to cooperate with nonlinear models. 

\textbf{The collaborative weight of two diverse representations help produce an accurate classifier}. Currently, more work is focusing on the neural network architecture and/or parameter tuning. However, the proposed DeepCWC neither pays much attention to deep learning model itself, nor tunes any parameters. Fusing multiple representations creates a robust classifier that works well on multiple deep learning models. The results are all at a state-of-the-art level.   

\textbf{The global average pool layer shows an effective capacity to extract discriminative features}. Global average pooling was proposed to enforce the learning of the class level feature maps \cite{Lin2013Network}. The experiments on layer analysis showed that using features extracted from the global average pool layer can produce a higher accuracy. This is true in the Inception and ResNet models, as shown in Fig. \ref{DeepLayers} (a) and (d), and Tab. \ref{table:StateOfArt}. That being said, it needs more computation due to the relatively larger size of the feature set from this layer.  

\textbf{Multiple layers in a deep CNN model show an effective capacity to extract discriminative features,} including the global average pooling layer \cite{Lin2013Network} and the fully connected layer. This is confirmed in the experiments, as shown in Fig. \ref{DeepLayers}, and Tab. \ref{table:StateOfArt}. However, the size of feature map decides the time consumption. 

\textbf{The proposed DeepCWC takes the NO. 1 position in current benchmark rank of Fashion-MNIST}. The most promising result is obtained on the VGG-16 model, which outperforms current leaders mainly using the WRN model.

\section{Conclusions}

We propose a deep collaborative weight-based classification (DeepCWC) method. It first performs the linear representation on original images and deep features, extracted from nonlinear neural networks. Then, both of them collaboratively weight each other to build a strong discriminative classifier. The method is extensively evaluated using multiple popular deep CNN models, like ResNet, Inception, and VGG. The experimental results are promising on more than one layers in these neural networks, with most of the results belonging to a state-of-the-art level.

The $l_2$-norm based CR model is chosen as the linear constraint in this work to enhance the classification based on pre-trained CNN models. However, there are still some questions, for example, whether there are other linear models (like sparse representation, dictionary learning, etc.), more suitable for the same task, or whether it can bring one more step of break-through when applied on locally trained and tuned CNN models. We will keep working on these open topics in the future.

\section{The Acknowledgements}

We gratefully acknowledge the support of NVIDIA Corporation with the donation of the Titan Xp GPU used for this research.

\bibliographystyle{abbrv}
\bibliography{DeepCWC}

\begin{thebibliography}{10}

\bibitem{Akhtar2017Efficient}
N.~Akhtar, F.~Shafait, and A.~Mian.
\newblock Efficient classification with sparsity augmented collaborative
  representation.
\newblock {\em Pattern Recognition}, 65:136--145, 2017.

\bibitem{Cai2016Probabilistic}
S.~Cai, L.~Zhang, W.~Zuo, and X.~Feng.
\newblock A probabilistic collaborative representation based approach for
  pattern classification.
\newblock In {\em Computer Vision and Pattern Recognition}, pages 2950--2959,
  2016.

\bibitem{OrlData}
A.~L. Cambridge.
\newblock The orl database of faces.
\newblock
  \url{http://www.cl.cam.ac.uk/research/dtg/attarchive/facedatabase.html}.
\newblock Online; accessed 12-Octobor-2017.

\bibitem{Chen2005FaceGT}
L.~Chen, H.~Man, and A.~V. Nefian.
\newblock Face recognition based on multi-class mapping of fisher scores.
\newblock {\em Pattern Recognition}, 38(6):799--811, 2005.

\bibitem{Deng2009Imagenet}
J.~Deng, W.~Dong, R.~Socher, L.-J. Li, K.~Li, and L.~Fei-Fei.
\newblock Imagenet: A large-scale hierarchical image database.
\newblock In {\em Computer Vision and Pattern Recognition, 2009. CVPR 2009.
  IEEE Conference on}, pages 248--255. IEEE, 2009.

\bibitem{Deng2013Defense}
W.~Deng, J.~Hu, and J.~Guo.
\newblock In defense of sparsity based face recognition.
\newblock In {\em Computer vision and pattern recognition (cvpr), 2013 ieee
  conference on}, pages 399--406. IEEE, 2013.

\bibitem{Georghiades2001FewYaleB}
A.~S. Georghiades, P.~N. Belhumeur, and D.~J. Kriegman.
\newblock From few to many: Illumination cone models for face recognition under
  variable lighting and pose.
\newblock {\em IEEE transactions on pattern analysis and machine intelligence},
  23(6):643--660, 2001.

\bibitem{He2016Deep}
K.~He, X.~Zhang, S.~Ren, and J.~Sun.
\newblock Deep residual learning for image recognition.
\newblock In {\em Proceedings of the IEEE conference on computer vision and
  pattern recognition}, pages 770--778. IEEE, 2016.

\bibitem{He2016Identity}
K.~He, X.~Zhang, S.~Ren, and J.~Sun.
\newblock Identity mappings in deep residual networks.
\newblock In {\em European Conference on Computer Vision}, pages 630--645.
  Springer, 2016.

\bibitem{Krizhevsky2009LearningCIFAR10}
A.~Krizhevsky and G.~Hinton.
\newblock Learning multiple layers of features from tiny images.
\newblock 2009.

\bibitem{Krizhevsky2012Imagenet}
A.~Krizhevsky, I.~Sutskever, and G.~E. Hinton.
\newblock Imagenet classification with deep convolutional neural networks.
\newblock In {\em Advances in neural information processing systems}, pages
  1097--1105, 2012.

\bibitem{LeCun2015Deep}
Y.~LeCun, Y.~Bengio, and G.~Hinton.
\newblock Deep learning.
\newblock {\em Nature}, 521(7553):436--444, 2015.

\bibitem{Li2018Decision}
Y.~Li, P.~Richtarik, L.~Ding, and X.~Gao.
\newblock On the decision boundary of deep neural networks.
\newblock {\em arXiv preprint arXiv:1808.05385}, 2018.

\bibitem{Lin2013Network}
M.~Lin, Q.~Chen, and S.~Yan.
\newblock Network in network.
\newblock {\em arXiv preprint arXiv:1312.4400}, 2013.

\bibitem{Martinez1998AR}
A.~M. Martinez.
\newblock The ar face database.
\newblock {\em CVC Technical Report}, 24, 1998.

\bibitem{Milborrow10MUCT}
S.~Milborrow, J.~Morkel, and F.~Nicolls.
\newblock {The MUCT Landmarked Face Database}.
\newblock {\em Pattern Recognition Association of South Africa}, 2010.
\newblock \url{http://www.milbo.org/muct}.

\bibitem{Nayar1996Columbia}
S.~Nayar, S.~Nene, and H.~Murase.
\newblock Columbia object image library (coil 100).
\newblock {\em Department of Comp. Science, Columbia University, Tech. Rep.
  CUCS-006-96}, 1996.

\bibitem{Peng2014Learning}
X.~Peng, L.~Zhang, Z.~Yi, and K.~K. Tan.
\newblock Learning locality-constrained collaborative representation for robust
  face recognition.
\newblock {\em Pattern Recognition}, 47(9):2794--2806, 2014.

\bibitem{Phillips2000Feret}
P.~J. Phillips, H.~Moon, S.~A. Rizvi, and P.~J. Rauss.
\newblock The feret evaluation methodology for face-recognition algorithms.
\newblock {\em IEEE Transactions on pattern analysis and machine intelligence},
  22(10):1090--1104, 2000.

\bibitem{Ristin2014Incremental}
M.~Ristin, M.~Guillaumin, J.~Gall, and L.~V. Gool.
\newblock Incremental learning of ncm forests for large-scale image
  classification.
\newblock In {\em Computer Vision and Pattern Recognition}, pages 3654--3661,
  2014.

\bibitem{Scardapane2017Group}
S.~Scardapane, D.~Comminiello, A.~Hussain, and A.~Uncini.
\newblock Group sparse regularization for deep neural networks.
\newblock {\em Neurocomputing}, 241:81--89, 2017.

\bibitem{Shao2017Performance}
L.~Shao, Z.~Cai, L.~Liu, and K.~Lu.
\newblock Performance evaluation of deep feature learning for rgb-d image/video
  classification.
\newblock {\em Information Sciences}, 385:266--283, 2017.

\bibitem{Sharif2014Cnn}
A.~Sharif~Razavian, H.~Azizpour, J.~Sullivan, and S.~Carlsson.
\newblock Cnn features off-the-shelf: an astounding baseline for recognition.
\newblock In {\em Proceedings of the IEEE conference on computer vision and
  pattern recognition workshops}, pages 806--813, 2014.

\bibitem{Simonyan2014Very}
K.~Simonyan and A.~Zisserman.
\newblock Very deep convolutional networks for large-scale image recognition.
\newblock {\em arXiv preprint arXiv:1409.1556}, 2014.

\bibitem{Szegedy2017Inception}
C.~Szegedy, S.~Ioffe, V.~Vanhoucke, and A.~A. Alemi.
\newblock Inception\-v4, inception\-resnet and the impact of residual
  connections on learning.
\newblock In {\em AAAI}, pages 4278--4284, 2017.

\bibitem{Valada2018Deep}
A.~Valada, L.~Spinello, and W.~Burgard.
\newblock Deep feature learning for acoustics-based terrain classification.
\newblock In {\em Robotics Research}, pages 21--37. Springer, 2018.

\bibitem{Wen2018Adaptive}
J.~Wen, B.~Zhang, Y.~Xu, J.~Yang, and N.~Han.
\newblock Adaptive weighted nonnegative low-rank representation.
\newblock {\em Pattern Recognition}, 81:326--340, 2018.

\bibitem{Wen2016Learning}
W.~Wen, C.~Wu, Y.~Wang, Y.~Chen, and H.~Li.
\newblock Learning structured sparsity in deep neural networks.
\newblock In {\em Advances in Neural Information Processing Systems}, pages
  2074--2082, 2016.

\bibitem{Wen2016Discriminative}
Y.~Wen, K.~Zhang, Z.~Li, and Y.~Qiao.
\newblock A discriminative feature learning approach for deep face recognition.
\newblock In {\em European Conference on Computer Vision}, pages 499--515.
  Springer, 2016.

\bibitem{Wright2009Robust}
J.~Wright, A.~Y. Yang, A.~Ganesh, S.~S. Sastry, and Y.~Ma.
\newblock Robust face recognition via sparse representation.
\newblock {\em Pattern Analysis and Machine Intelligence, IEEE Transactions
  on}, 31(2):210--227, 2009.

\bibitem{Xiao2017Fashion}
H.~Xiao, K.~Rasul, and R.~Vollgraf.
\newblock Fashion-mnist: a novel image dataset for benchmarking machine
  learning algorithms.
\newblock {\em arXiv preprint arXiv:1708.07747}, 2017.

\bibitem{Xiao2016Learning}
T.~Xiao, H.~Li, W.~Ouyang, and X.~Wang.
\newblock Learning deep feature representations with domain guided dropout for
  person re-identification.
\newblock In {\em Proceedings of the IEEE Conference on Computer Vision and
  Pattern Recognition}, pages 1249--1258, 2016.

\bibitem{Xiao2014Error}
T.~Xiao, J.~Zhang, K.~Yang, Y.~Peng, and Z.~Zhang.
\newblock Error-driven incremental learning in deep convolutional neural
  network for large-scale image classification.
\newblock In {\em ACM International Conference on Multimedia}, pages 177--186,
  2014.

\bibitem{Xin2016Maximal}
B.~Xin, Y.~Wang, W.~Gao, D.~Wipf, and B.~Wang.
\newblock Maximal sparsity with deep networks?
\newblock In {\em Advances in Neural Information Processing Systems}, pages
  4340--4348, 2016.

\bibitem{Xu2017Sample}
Y.~Xu, Z.~Li, B.~Zhang, J.~Yang, and J.~You.
\newblock Sample diversity, representation effectiveness and robust dictionary
  learning for face recognition.
\newblock {\em Information Sciences}, 375(C):171--182, 2017.

\bibitem{Xu2011Two}
Y.~Xu, D.~Zhang, J.~Yang, and J.-Y. Yang.
\newblock A two-phase test sample sparse representation method for use with
  face recognition.
\newblock {\em IEEE Transactions on Circuits and Systems for Video Technology},
  21(9):1255--1262, 2011.

\bibitem{Xu2017New}
Y.~Xu, Z.~Zhong, J.~Yang, J.~You, and D.~Zhang.
\newblock A new discriminative sparse representation method for robust face
  recognition via $ l\_ $\{$2$\}$ $ regularization.
\newblock {\em IEEE transactions on neural networks and learning systems},
  28(10):2233--2242, 2017.

\bibitem{Yu2012Large}
H.~F. Yu, C.~J. Hsieh, K.~W. Chang, and C.~J. Lin.
\newblock Large linear classification when data cannot fit in memory.
\newblock {\em ACM Transactions on Knowledge Discovery from Data (TKDD)},
  5(4):23, 2012.

\bibitem{Zagoruyko2016Wide}
S.~Zagoruyko and N.~Komodakis.
\newblock Wide residual networks.
\newblock {\em arXiv preprint arXiv:1605.07146}, 2016.

\bibitem{Zeng2017Improving}
S.~Zeng, J.~Gou, and X.~Yang.
\newblock Improving sparsity of coefficients for robust sparse and
  collaborative representation-based image classification.
\newblock {\em Neural Computing \& Applications}, pages 1--14, 2017.

\bibitem{Zeng2017Multiplication}
S.~Zeng, X.~Yang, and J.~Gou.
\newblock Multiplication fusion of sparse and collaborative representation for
  robust face recognition.
\newblock {\em Multimedia Tools and Applications}, 76(20):20889--20907, 2017.

\bibitem{Zeng2017Joint}
S.~Zeng, B.~Zhang, and Y.~Du.
\newblock Joint distances by sparse representation and locality-constrained
  dictionary learning for robust leaf recognition.
\newblock {\em Computers and Electronics in Agriculture}, 142:563--571, 2017.

\bibitem{Zeng2018Robust}
S.~Zeng, B.~Zhang, Y.~Lan, and J.~Gou.
\newblock Robust collaborative representation-based classification via
  regularization of truncated total least squares.
\newblock {\em Neural Computing and Applications}, pages 1--9, 2018.

\bibitem{Zhang2012Sparse}
L.~Zhang, M.~Yang, and X.~Feng.
\newblock Sparse representation or collaborative representation: Which helps
  face recognition?
\newblock In {\em IEEE International Conference on Computer Vision}, pages
  471--478, 2012.

\bibitem{Zhong2017Random}
Z.~Zhong, L.~Zheng, G.~Kang, S.~Li, and Y.~Yang.
\newblock Random erasing data augmentation.
\newblock {\em arXiv preprint arXiv:1708.04896}, 2017.

\bibitem{Zhou2014Learning}
B.~Zhou, A.~Lapedriza, J.~Xiao, A.~Torralba, and A.~Oliva.
\newblock Learning deep features for scene recognition using places database.
\newblock In {\em Advances in neural information processing systems}, pages
  487--495, 2014.

\end{thebibliography}
\end{document}